\documentclass[letterpaper, 10pt, conference]{ieeeconf}      

\IEEEoverridecommandlockouts                              

\usepackage{amsmath,amssymb} \usepackage{url}
\usepackage{xcolor}
\usepackage{nccmath} 
\usepackage{float}
\usepackage{graphicx} 
\usepackage{gensymb} 
\usepackage{CJKutf8}
\usepackage[ruled,vlined,linesnumbered]{algorithm2e} 
\usepackage{color}
\usepackage{booktabs}
\usepackage{balance}
\usepackage{url}
\usepackage[utf8]{inputenc} 
\usepackage[T1]{fontenc}    
\usepackage{hyperref}       
\usepackage{url}            
\usepackage{booktabs}       
\usepackage{amsfonts}       
\usepackage{nicefrac}       
\usepackage{microtype}      
\usepackage{xcolor}   
\usepackage{siunitx}
\usepackage{mathtools}

\usepackage{amsmath}
\usepackage{amsfonts}

\newcommand{\Sspace}{\mathcal{S}}         
\newcommand{\Aspace}{\mathcal{A}}         
\newcommand{\Prob}{P}               
\newcommand{\policy}{\pi}           
\newcommand{\states}{s}             
\newcommand{\action}{a}           
\newcommand{\Expect}{\mathbb{E}}    
\newcommand{\ProbMeasure}{\mathbb{P}} 

\newcommand{\Ssafe}{\mathcal{S}_{\text{safe}}}     
\newcommand{\Sunsafe}{\mathcal{S}_{\text{unsafe}}}   

\newcommand{\SIC}{\mathcal{S}_{IC}}                 

\newcommand{\policySafe}{\Pi_{\text{safe}}}             
\newcommand{\policyC}{\Pi_{C}}                   
\newcommand{\policyUC}{\Pi_{UC}}                 

\newcommand{\R}{\mathbb{R}}                

\newcommand{\state}{x}                    
\newcommand{\control}{u}                  

\newcommand{\AL}{A_L}                     
\newcommand{\BL}{B_L}                     

\newcommand{\gammaCBF}{\gamma_{\text{CBF}}}

\newcommand{\gammaRL}{\gamma_{\text{RL}}}  

\title{\LARGE \bf
End-to-End Humanoid Robot Safe and Comfortable Locomotion Policy
}

\author{Zifan Wang$^{*1}$, Xun Yang$^{*1}$, Jianzhuang Zhao$^{3}$, Jiaming Zhou$^{1}$, Teli Ma$^{1}$, Ziyao Gao$^{1}$, 
\\ Arash Ajoudani$^{3}$, Junwei Liang$^{\dag1,2}$ 
\thanks{*Equal contribution. \dag Corresponding author. $^{1}$The Hong Kong University of Science and Technology (Guangzhou), $^{2}$The Hong Kong University of Science and Technology,  $^{3}$Human-Robot Interfaces and Interaction Lab., Istituto Italiano di Tecnologia, Italy.}}

\begin{document}

\maketitle
\thispagestyle{empty}
\pagestyle{empty}

\begin{abstract}
The deployment of humanoid robots in unstructured, human-centric environments requires navigation capabilities that extend beyond simple locomotion to include robust perception, provable safety, and socially aware behavior. Current reinforcement learning approaches are often limited by blind controllers that lack environmental awareness or by vision-based systems that fail to perceive complex 3D obstacles. In this work, we present an end-to-end locomotion policy that directly maps raw, spatio-temporal LiDAR point clouds to motor commands, enabling robust navigation in cluttered dynamic scenes. We formulate the control problem as a Constrained Markov Decision Process (CMDP) to formally separate safety from task objectives. Our key contribution is a novel methodology that translates the principles of Control Barrier Functions (CBFs) into costs within the CMDP, allowing a model-free Penalized Proximal Policy Optimization (P3O)—to enforce safety constraints during training. Furthermore, we introduce a set of comfort-oriented rewards, grounded in human-robot interaction research, to promote motions that are smooth, predictable, and less intrusive. We demonstrate the efficacy of our framework through a successful sim-to-real transfer to a physical humanoid robot, which exhibits agile and safe navigation around both static and dynamic 3D obstacles. Project Page:\url{https://github.com/aCodeDog/SafeHumanoidsPolicy}
\\
\\
\textbf{Keywords:} Humanoid Robot, Locomotion, Reinforcement Learning, Collision Avoidance, LiDAR Perception
\end{abstract}

\section{INTRODUCTION}
\label{sec:intro}
The vision of humanoid robots seamlessly coexisting and collaborating with people in everyday environments presents a grand challenge for robotics. A fundamental prerequisite for this vision is the ability to navigate complex, human-centric spaces safely and efficiently. This requires more than just dynamic locomotion; it demands a holistic integration of 3D perception, principled safety, and social awareness, an area where current controllers often fall short.

A significant body of state-of-the-art research in legged locomotion has relied on reinforcement learning (RL) to develop controllers that are "blind," using only proprioceptive feedback~\cite{margolis2022walktheseways,margolisyang2022rapid,hwangbo2019learning}. While these methods have achieved remarkable agility on flat or moderately uneven terrain, they are inherently incapable of navigating environments with obstacles.

To overcome this, recent work has incorporated exteroceptive sensing, primarily using depth cameras to generate 2D height maps for local terrain awareness\cite{cheng2024extreme,zhuang2024humanoid,zhuang2023robot,lee2024learning}.Depth cameras are sensitive to lighting conditions and have a limited field of view\cite{langmann2012depth,song2016robust,zhong2021survey}.And the reduction of 3D sensory information to a 2D elevation map makes the robot blind to any non-ground-level obstacles, such as overhanging clutter or the upper bodies of other agents\cite{miki2022elevation,fankhauser2014robot,souza2016occupancy}. This limitation poses a substantial collision risk for a full-body humanoid robot. We argue that for robust navigation in cluttered indoor spaces, a perception modality that is lighting-invariant and provides direct 3D information is essential. LiDAR sensors meet these criteria, yet their integration into end-to-end locomotion policies remains limited~\cite{wang2025omni,he2024agile}.

Even with robust perception, ensuring safety is a non-trivial challenge. The common approach of shaping reward functions with penalties for collisions is often brittle, difficult to tune, and can lead to overly conservative or still-unsafe behaviors~\cite{fu2022coupling,lee2024learning,bohorquez2016safe}. Furthermore, for robots intended to operate alongside people, merely avoiding collisions is insufficient. The robot's motion must also be psychologically comfortable—that is, predictable, fluid, and non-threatening—to foster trust and acceptance~\cite{mitsunaga2008adapting,rodriguez2015bellboy,wang2019learning,becker2023customer}. This higher-level, human-centric aspect of motion planning is rarely considered in locomotion policies.

In this work, we address these gaps with an integrated, end-to-end framework. Our contributions are:
\begin{enumerate}
    \item \textbf{A LiDAR-driven end-to-end policy} that processes raw 3D point clouds to navigate complex environments, overcoming the limitations of blind and 2D-vision-based approaches.
    \item \textbf{A principled safety framework} based on Constrained Reinforcement Learning (CMDP). We introduce a novel method to translate model-based Control Barrier Function (CBF) principles into costs for a model-free RL algorithm, P3O, enabling robust safety enforcement.
    \item \textbf{A comfort-oriented reward structure} explicitly designed to produce socially aware motions by penalizing behaviors known to cause human discomfort, such as high approach speeds and unpredictable movements, as identified in HRI research.
    \item \textbf{Successful real-world deployment} on a humanoid robot, demonstrating agile and robust avoidance of diverse static and dynamic 3D obstacles in cluttered environments.
\end{enumerate}

\section{RELATED WORK}
\label{sec:relatedwork}

\subsection{Legged Robot Locomotion and Perception}
 RL has produced policies capable of remarkable dynamic skills, yet many still lack the environmental perception needed for navigation in cluttered spaces\cite{hoeller2024anymal, zhuang2024humanoid,wang2024arm}. Recent work has begun to incorporate exteroceptive sensing. Policies using height maps derived from depth cameras or LiDAR have successfully traversed uneven terrain\cite{lee2024learning,cheng2024extreme,chanesane2024solo,NavRL,wang2025omni}.

\subsection{Safe Reinforcement Learning}
Ensuring safety during RL is critical for real-world deployment\cite{brunke2022safe,gu2022review}. A common but often unreliable method is to use negative rewards to penalize unsafe actions. A more structured approach is to formulate the problem as a CMDP, which separates the task objective (reward) from safety specifications (constraints)\cite{zhang2022penalized,wachi2020safe}. This allows for the use of specialized algorithms that aim to satisfy constraints while maximizing rewards.

Within this domain, Control-Theoretic Methods offer strong safety guarantees. CBFs define a safe region of the state space and can be used to synthesize controllers that are guaranteed to remain within it\cite{desai2022clf,manjunath2021safe}. However, classical CBF approaches typically require an accurate analytical model of the system dynamics, making them difficult to apply directly in model-free RL.

Our work bridges this gap. We draw inspiration from CBF theory to define a safety condition but implement it as a cost function within a CMDP inspired by~\cite{choi2020reinforcement,emam2022safe}. This allows us to use a powerful, model-free constrained policy optimization algorithm, P3O~\cite{zhang2022penalized}, to learn a safe policy. P3O is a first-order method, making it more computationally efficient than higher-order alternatives like CPO , and its use of normalized advantage functions improves training stability. By integrating CBF principles into a proven, practical CMDP algorithm, we develop a robust framework for learning safe, perception-driven locomotion.

\section{Definitions}
\newtheorem{definition}{Definition}
\newtheorem{proposition}{Proposition} 

\subsection{Constrained Markov Decision Process}

The task of learning a safe locomotion policy is formulated as a CMDP which is defined by a tuple $(\Sspace, \Aspace, P, R, \{C_j\}, \{\epsilon_j\}, \gammaRL)$, where:
\begin{itemize}
    \item $\Sspace$ is the set of states $\states_k$.
    \item $\Aspace$ is the set of actions $\action_k$ (equivalent to $\control_k$).
    \item $P(\states_{k+1} | \states_k, \action_k)$ is the state transition probability.
    \item $R(\states_k, \action_k, \states_{k+1})$ is the reward function.
    \item $C_j(\states_k, \action_k, \states_{k+1})$ is the $j$-th instantaneous cost function.
    \item $\epsilon_j$ is the threshold for the $j$-th constraint.
    \item $\gammaRL \in [0,1)$ is the discount factor for reinforcement learning.
\end{itemize}
The objective is to find a policy $\pi: \mathcal{S} \rightarrow \mathcal{P(A)}$ that maximizes the expected discounted sum of rewards, $J_R(\pi) = \mathbb{E}_{\tau \sim \pi} [\sum_{t=0}^{\infty} \gamma^t R(s_t, a_t, s_{t+1})]$, subject to constraints on the expected discounted sum of costs:
\begin{equation}\label{eq:represent}
\begin{gathered}
 J_{C_j}(\pi) = \mathbb{E}_{\tau \sim \pi} \left[\sum_{t=0}^{\infty} \gamma^t C_j(s_t, a_t, s_{t+1})\right] \le \epsilon_j, \\ 
\quad \forall j \in \{1, ..., m\}. 
\end{gathered}
\end{equation}

 We partition the state space $\Sspace$ into a set of \textbf{safe states}, denoted $\Ssafe$, and a set of \textbf{unsafe states}, denoted $\Sunsafe$, such that $\Sspace = \Ssafe \cup \Sunsafe$ and $\Ssafe \cap \Sunsafe = \emptyset$. The set $\Ssafe$ represents configurations where the system can operate without risk, while $\Sunsafe$ represents configurations that must be avoided. We introduce the following definitions related to human comfort, assuming operation primarily within the safe space $\Ssafe$.

\begin{figure}[h]
    \centering
    \vspace{-6pt}
    \includegraphics[width = 0.48\textwidth]{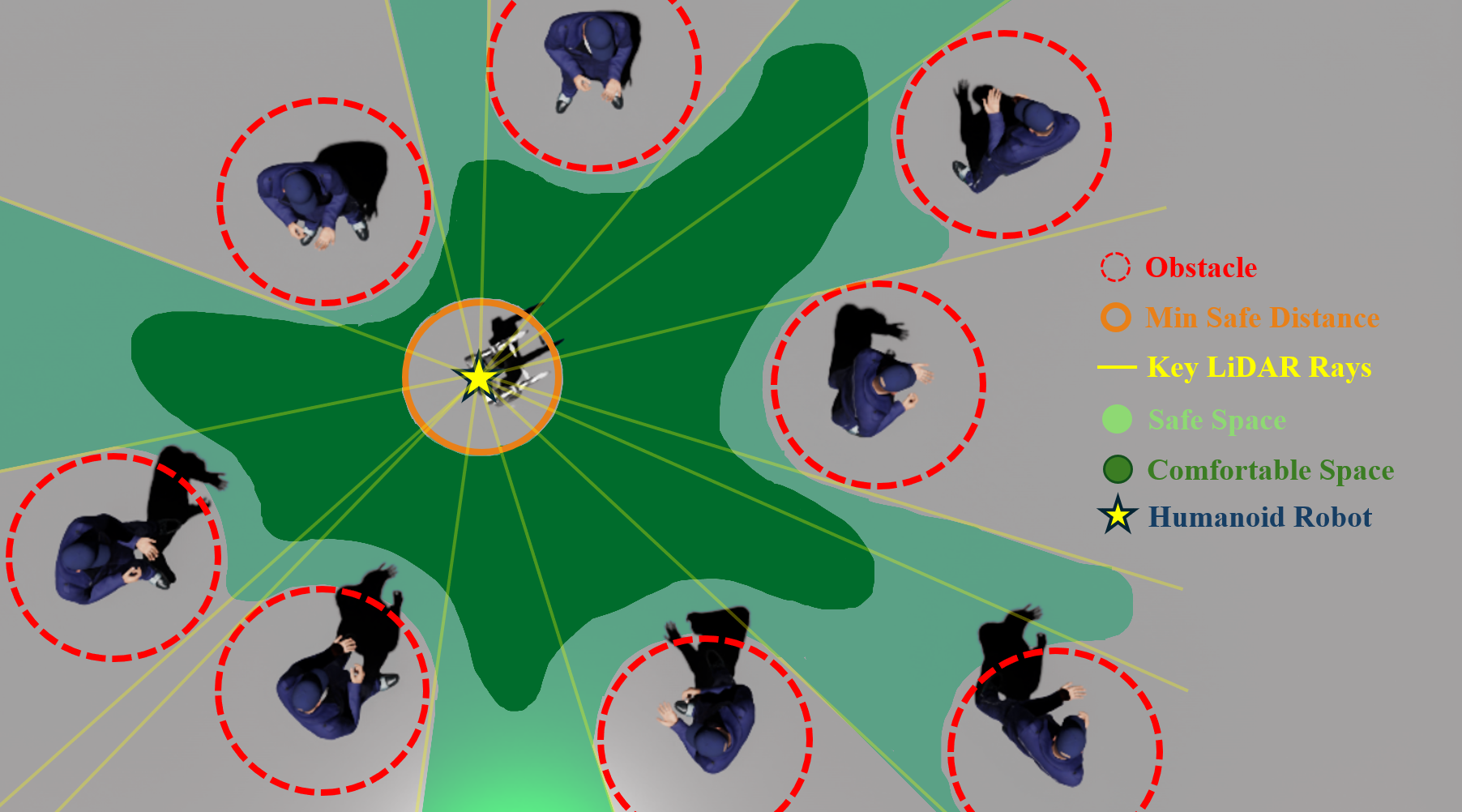}
    \caption{Illustration of safe and comfortable space}
    \label{fig:area}
    \vspace{-6pt}
\end{figure}

\begin{definition}[Interactive Comfortable Space]
\label{def:ics}
Given the set of safe states $\Ssafe$ in a CMDP, the \textbf{Interactive Comfortable Space} (ICS), denoted by $\SIC$, is a subset of the safe state space, $\SIC \subseteq \Ssafe$. This subspace represents state configurations wherein the agent's (e.g., robot's) presence and subsequent actions, governed by a comfortable policy (see Definition \ref{def:policies}), are intended to avoid causing disruption or discomfort to proximate passive agents (e.g., humans, animals, see Fig. \ref{fig:area}) sharing the operational environment. The specific definition of $\SIC$ is based on proximity, velocity, and agent's state.
\end{definition}
\vspace{-6pt}
\vspace{-6pt}
\begin{definition}[Safe and Comfortable Policies]
\label{def:policies}
Consider a CMDP $\hat{M}$ with safe states $\Ssafe$, unsafe states $\Sunsafe$, and Interactive Comfortable Space $\SIC \subseteq \Ssafe$. Let a trajectory starting from $\states_0$ under a stochastic policy $\policy(\action|\states)$ be denoted by $\{\states_t\}_{t \ge 0}$, where $\states_{t+1} \sim \Prob(\states_{t+1}|\states_t, \action_t)$ and $\action_t \sim \policy(\action_t|\states_t)$. We define classes of policies based on their safety and comfort properties:
\begin{itemize}
    \item A policy $\policy$ is \textbf{safe} if, for any initial state $\states_0 \in \Ssafe$, the resulting trajectory satisfies $\ProbMeasure(\states_t \in \Ssafe, \forall t \ge 0 \mid \states_0, \policy) = 1$. Let $\policySafe$ denote the set of all safe policies.
    \item A policy $\policy$ is \textbf{comfortable} if it is safe ($\policy \in \policySafe$) and additionally, for any initial state $\states_0 \in \SIC$, the resulting trajectory satisfies $\ProbMeasure(\states_t \in \SIC, \forall t \ge 0 \mid \states_0, \policy) = 1$. Let $\policyC$ denote the set of all comfortable policies. By definition, $\policyC \subseteq \policySafe$.
    \item A policy $\policy$ is \textbf{uncomfortable but safe} if $\policy \in \policySafe$ but $\policy \notin \policyC$. This implies that the policy guarantees safety (stays within $\Ssafe$) but does not guarantee remaining within the comfortable subspace $\SIC$ when starting from $\SIC$. Let $\policyUC = \policySafe \setminus \policyC$.
    \item A \textbf{unsafe policy} $\policy \notin \policySafe$ means there exists some initial state $\states_0 \in \Ssafe$
    such that $\ProbMeasure(\exists t > 0 \text{ s.t. } \states_t \in \Sunsafe \mid \states_0, \policy) > 0$. 
    Let  $ \Pi_{\text{unsafe}} $ be the set of all unsafe policies.
\end{itemize}
\end{definition}

\begin{definition}[Comfortable Target Tracking and Obstacle Avoidance]
\label{def:perfect_objective}
Given a CMDP $\hat{M}$, a target state $\states_{\text{goal}}$ or target region $\mathcal{S}_{\text{goal}}$, the safe state space $\Ssafe$, and the Interactive Comfortable Space $\SIC$, the objective of \textbf{Comfortable Target Tracking and Obstacle Avoidance} is to find an optimal comfortable policy $\policy^{*}_{C} \in \policyC$ that maximizes a task-specific performance criterion $\mathcal{J}(\policy)$ (e.g., maximizes rewards $\states_{\text{goal}}$, minimizes expected cost to reach $\mathcal{S}_{\text{goal}}$) subject to the safety and comfort constraints defined in Definition \ref{def:policies}.
Specifically, $\policy^{*}_{C} = \arg\max_{\policy \in \policyC} \mathcal{J}(\policy)$.
\end{definition}

\subsection{System Dynamics}
We consider a robot operating in discrete time steps $k$. Its state at time step $k$ is denoted by $\state_k \in \Sspace \subseteq \R^n$, and the control input applied is $\control_k \in \Aspace \subseteq \R^m$. The system dynamics are described by:
\begin{equation}
    \state_{k+1} = f(\state_k, \control_k)
    \label{eq:general_dynamics}
\end{equation}
We assume the system dynamics is linear:
\begin{equation}
    \state_{k+1} = \AL \state_k + \BL \control_k
    \label{eq:lip_dynamics}
\end{equation}
where $\AL$ is the state‐transition matrix, and $\BL$ is the input matrix.

\subsection{Discrete Control Barrier Functions (DCBF)}
A DCBF $h(\state_k): \R^n \rightarrow \R$ is used to define a safe set $\mathcal{C} = \{\state_k \in \Sspace | h(\state_k) \ge 0\}$~\cite{agrawal2017discrete}. A control input $\control_k$ renders the set $\mathcal{C}$ forward invariant if for some $0 < \gammaCBF \le 1$, the following condition holds  :
\begin{equation}
    h(\state_{k+1}) + (\gammaCBF - 1)h(\state_k) \ge 0
    \label{eq:dcbf_condition}
\end{equation}
This ensures that if $\state_k \in \mathcal{C}$, then $\state_{k+1}$ also remains in (a subset of) $\mathcal{C}$.

\section{METHODOLOGY}

Our framework is designed to learn a  \textit{Comfortable Policy} ($\pi_C^*$) that operates within the formally defined safe and comfortable state spaces. The methodology integrates perception, safety constraints, and policy learning into a single end-to-end model. We translate a CBF-based safety condition into a cost function for our CMDP, which is then solved using the P3O algorithm to find a policy that maximizes the task objective while satisfying all constraints.

\begin{figure}[thpb]
    \centering
    \includegraphics[width=1.0\columnwidth]{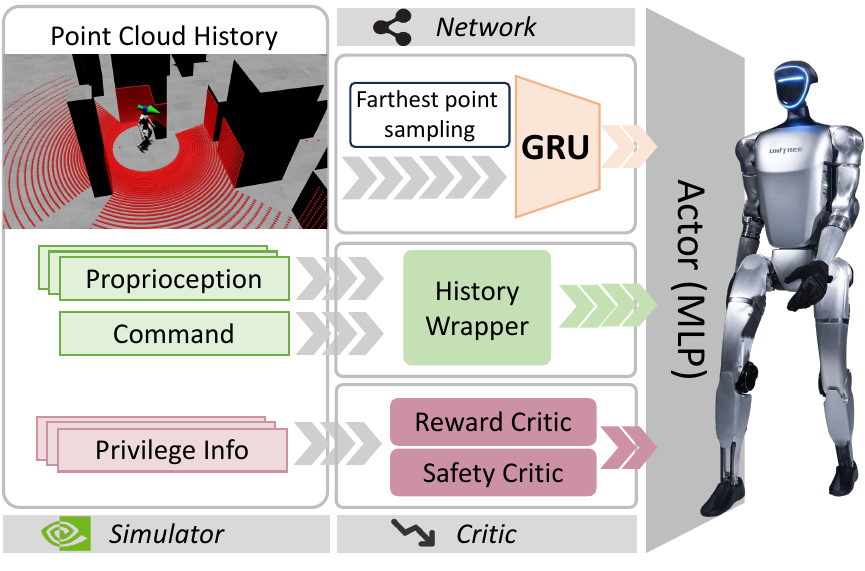} 
    \caption{Overview of the proposed training framework. Raw sensor data is processed by the Actor (policy).}
    \label{fig:framework}
    \vspace{-24pt}
\end{figure}

\subsection{Network Inputs and Architecture}
\subsubsection{Actor Inputs (Policy)}
\begin{itemize}
    \item \textbf{Proprioceptive and Command History:} To capture the robot's recent motion and intentions, the polcy is provided with a history of the 10 timesteps. Each step in this history contains the robot's joint positions, joint velocities, joint accelerations, the previous action taken, base linear velocity, base angular velocity, the gravity vector in the base frame, base height, and the user command (linear x/y velocity, angular yaw velocity).
    \item \textbf{Exteroceptive (LiDAR) Features:} The environment is perceived through a feature vector extracted from the raw LiDAR point cloud. This LiDAR embedding, which captures the essential geometric information of the surroundings, is a 64-dimensional vector.
\end{itemize}

\subsubsection{Critic Inputs (Privileged Information)}
During training, the reward and safety critics receive all the information available to the actor, plus additional privileged information that is only available in simulation. The privileged information includes the true distance and velocity of the nearest obstacle in 8 discrete directions around the robot,the contact force of whole body link, and safe condition(the joints limits, minimum safe distance) . 

\subsubsection{Network Architecture}
As illustrated in Figure \ref{fig:framework}, the LiDAR feature history is processed by a Gated Recurrent Unit (GRU) to extract temporal patterns. The output of the GRU is then concatenated with the flattened proprioceptive and command history. This combined feature vector is passed through a series of fully connected layers forming a Multi-Layer Perceptron (MLP), which constitutes the main body of the actor and critic networks.

\subsection{Enforcing the Safe State Space via LDCBF Cost}
To ensure the learned policy $\pi$ is a member of the set of \textit{Safe Policies} ($\Pi_{safe}$), we must prevent it from entering the unsafe state space $\mathcal{S}_{unsafe}$. We define the boundary of the \textit{Safe State Space} ($\mathcal{S}_{safe}$) by enforcing a minimum distance $D_{min}$ from obstacles, formulated using a Linear Discrete-Time Control Barrier Function (LDCBF).

Assuming the obstacle boundary is locally approximated by a hyperplane, the LDCBF barrier function $h_D(\boldsymbol{s}_k)$ is defined as the signed distance from the robot's position $\boldsymbol{p}(\boldsymbol{s}_k)$ to a safety margin offset from the obstacle:
\begin{equation}
    h_D(\boldsymbol{s}_k) = \boldsymbol{\eta}_k^T (\boldsymbol{p}(\boldsymbol{s}_k) - \boldsymbol{o}_k) - D_{min}
\end{equation}
where $\boldsymbol{o}_k$ is the closest point on the obstacle and $\boldsymbol{\eta}_k$ is the outward-pointing normal vector, both derived from the current LiDAR scan. Safety requires $h_D(\boldsymbol{s}_k) \ge 0$.

For a linear system model $\boldsymbol{s}_{k+1} = \boldsymbol{A}_L \boldsymbol{s}_k + \boldsymbol{B}_L \boldsymbol{u}_k$, the one-step-ahead safety condition $h_D(\boldsymbol{s}_{k+1}) \ge (1-\gamma_{CBF})h_D(\boldsymbol{s}_k)$ can be written as a linear constraint on the control input $\boldsymbol{u}_k$:
\begin{equation}
    G_D(\boldsymbol{s}_k, \boldsymbol{u}_k) \ge 0
    \label{eq:ldcbf_constraint}
\end{equation}
where $G_D$ is a function affine in $\boldsymbol{u}_k$. Since our framework is model-free, we do not use this constraint to filter actions directly. Instead, we transform its violation into an instantaneous cost function for the CMDP:
\begin{equation}
    C_{D}(\boldsymbol{s}_k, \boldsymbol{u}_k) = \max\{0, -G_D(\boldsymbol{s}_k, \boldsymbol{u}_k)\}
\end{equation}
This cost is positive only if the chosen action $\boldsymbol{u}_k$ is predicted to violate the safety barrier. 

\subsection{Learning Comfortable Policies via Rewards and Costs}
The policy is trained within a CMDP framework where the reward function guides the agent toward completing its task in a comfortable and socially acceptable manner, while the cost functions define hard safety and operational constraints. This design is informed by established principles in Human-Robot Interaction (HRI) to enhance perceived safety and comfort.

\subsubsection{Reward Function}
The total reward $\mathcal{R}_k$ is a weighted sum of task-oriented and comfort-oriented components. Research in HRI consistently demonstrates that human comfort is influenced by a robot's speed, proximity, ad the predictability of its movements. Abrupt or head-on movements are perceived as more threatening than smooth, tangential ones. Our reward structure is designed to directly incorporate these findings.

\begin{table}[h!]
\centering
\caption{Unified Reward and Cost Function Components}
\label{tab:rewards_and_costs}
\resizebox{\columnwidth}{!}{%
\begin{tabular}{@{}llc@{}}
\toprule
\textbf{Component} & \textbf{Equation / Notation} & \textbf{Weight} \\ \midrule
\multicolumn{3}{l}{\textit{\textbf{Task-Oriented Rewards}}} \\
Velocity Tracking & $\exp(-\alpha_v \|\mathbf{v}_k - \mathbf{v}_k^{\text{cmd}}\|^2)$ & 2.0 \\
Yaw Rate Tracking & $\exp(-\alpha_\omega (\omega_{z,k} - \omega_{z,k}^{\text{cmd}})^2)$ & 0.5 \\
\midrule
\multicolumn{3}{l}{\textit{\textbf{Auxiliary Rewards}}} \\
    Z velocity & \(v_{z}^2\) & \num{-3e-4} \\
    Link Collision &\(||Force^{\text{PenltyLink}}_{xy}||^2\)  & $-0.02$ \\  %
    Joint Torques & \(||\boldsymbol{\tau}||^2\) & \num{-1e-6} \\ 
    Joint Velocities & \(||\dot{\mathbf{q}}||^2\) & \num{-1e-6} \\ 
    Joint Accelerations & \(||\ddot{\mathbf{q}}||^2\) & \num{-2.5e-7} \\ 
    Action Smoothing & \(||\mathbf{a}_{t-1} - \mathbf{a}_t||^2\) & \num{-5e-3} \\ 
    Action Smoothing rate & \(||\mathbf{a}_{t-2} - 2\mathbf{a}_{t-1} + \mathbf{a}_t||^2\) & \num{-1e-5} \\
\midrule
\multicolumn{3}{l}{\textit{\textbf{Comfort-Oriented Rewards}}} \\
Proxemic Comfort & $\exp(-\alpha_p (d_{\text{human}, k} - d_{\text{social}})^2)$ & 1.5 \\
Safe Approach Velocity & $- \max(0, -\mathbf{v}_k \cdot \boldsymbol{\eta}_k)$ & -1.0 \\
Safe Approach Acceleration & $- \max(0, -\mathbf{a}_k \cdot \boldsymbol{\eta}_k)$ & -1.0 \\
Tangential Avoidance & $1 - \max(0, \hat{\mathbf{v}}_k \cdot (-\hat{\mathbf{d}}_{\text{obs},k}))$ & 1.0 \\

\midrule
\multicolumn{3}{l}{\textit{\textbf{Constraints}}}\\
Safety Distance Violation & $C_{\text{safe}} = \mathbf{1}_{D_{\text{obs},k} < d_{\text{safe}}}$ & $d_j = 0.0$ \\
Joint Limit Violation & $C_{q} = \sum_i \mathbf{1}_{q_{k,i} > q_{i}^{\text{max}} \| q_{k,i} < q_{i}^{\text{min}}}$ & $d_j = 0.0$ \\
Self-Collision & $C_{\text{coll}} = \mathbf{1}_{\text{links intersect}}$ & $d_j = 0.0$ \\
\bottomrule
\end{tabular}%
}
\vspace{-10pt}
\end{table}

The \textbf{Task-Oriented Rewards} (Table \ref{tab:rewards_and_costs}) provide the primary objective for the robot to follow commanded velocities. The core of our socially-aware behavior is shaped by the \textbf{Comfort-Oriented Rewards} (Table \ref{tab:rewards_and_costs}).

Based on studies of proxemics~\cite{app9235152,RUBAGOTTI2022104047,takayama2009influences,6899348,kim2014social}, we introduce a \textit{Proxemic Comfort} reward, where $d_{\text{social}}$ is the ideal social distance of 1.2 meters. This reward encourages the robot to maintain this distance from a person ($d_{\text{human}}$), peaking at the desired distance and decreasing as the robot gets either closer or farther away. To ensure motions are perceived as non-threatening, we penalize the dynamics of approach towards any obstacle (including people), a concept supported by research on speed and separation monitoring. The \textit{Safe Approach Velocity} and \textit{Safe Approach Accel} terms penalize the components of velocity $\mathbf{v}_k$ and acceleration $\mathbf{a}_k$ that are normal to the nearest obstacle surface (where $\boldsymbol{\eta}_k$ is the surface normal vector pointing away from the obstacle). This discourages the robot from moving directly and rapidly towards obstacles. Complementing this, the \textit{Tangential Avoidance} reward encourages the robot's velocity vector $\hat{\mathbf{v}}_k$ to be perpendicular to the direction of the nearest obstacle $\hat{\mathbf{d}}_{\text{obs},k}$, promoting smoother, arcing avoidance maneuvers rather than abrupt stops. Finally, \textit{Motion Smoothness} terms penalize high joint accelerations to ensure overall motion fluency, a key factor in perceived safety~.

\subsubsection{Cost Functions}
Costs define the hard safety boundaries of $\mathcal{S}_{unsafe}$ and are used for the CMDP constraints. These represent conditions that must be strictly avoided.

As shown in Table \ref{tab:rewards_and_costs}, the \textit{Safety Distance Violation} cost is a binary penalty triggered if the distance to any obstacle $D_{\text{obs},k}$ falls below the hard safety margin $d_{\text{safe}} = 0.8$ meters. This forms the primary definition of the unsafe state space, $\mathcal{S}_{unsafe}$. Additional costs for violating joint and torque limits, as well as for self-collision, are included to ensure the physical integrity of the robot. These costs are handled by the P3O algorithm to stringently enforce safe operation.
\subsection{Training with P3O}
To find an optimal policy that maximizes the task reward subject to the safety and physical cost constraints, we employ the Normalized Penalized Proximal Policy Optimization (P3O) algorithm. P3O is a first-order constrained RL method recognized for its stability and practical effectiveness in robotics. It augments the PPO objective with a penalty term for each constraint violation:
\begin{equation}
    L^{P3O}(\theta') = L_{R}^{CLIP,N}(\theta') - \sum_j \kappa_j \cdot \max\{0, L_{C_j}^{\text{VIOL,N}}(\theta')\}
    \label{eq:p3o_objective}
\end{equation}
Here, $\kappa_j$ is a tunable hyperparameter weighting the penalty for the $j$-th constraint. The term $L_{R}^{CLIP,N}$ is the standard PPO clipped objective using normalized reward advantages. The violation term for each cost, $L_{C_j}^{\text{VIOL,N}}$, is what connects the instantaneous cost $C_j$ to the policy update:
\begin{equation}
    L_{C_j}^{\text{VIOL,N}}(\theta') = L_{C_j}^{CLIP,N}(\theta') + \frac{(1-\gamma)(J_{C_j}(\pi_{\theta})-d_j)+\mu_{C_j}}{\sigma_{C_j}}
\end{equation}
where $J_{C_j}(\pi_{\theta}) = \Expect[\sum_{k=0}^\infty \gamma^k C_{j,k}]$ is the expected cumulative cost, and $L_{C_j}^{CLIP,N}$ is its clipped advantage estimate. $\mu_{C_j}$ and $\sigma_{C_j}$ are the mean and standard deviation of the un-normalized cost advantages calculated from the batch of samples for the $j$-th constraint. $d_j$ is the predefined threshold for the $j$-th cost constraint. $\gamma$ is the discount factor.



\section{EXPERIMENTS}
\label{sec:experiments}
We conduct a series of experiments in both high-fidelity simulation and on the physical Unitree G1 humanoid robot to validate the proposed end-to-end locomotion policy. Our evaluations are designed to test the policy's effectiveness in three key areas: robust navigation in complex 3D environments, principled safety enforcement, and the generation of comfortable, socially aware motions.
\subsection{Experimental Setup}
\begin{itemize}
    \item \textbf{Robot Platform:} The experiments utilize the Unitree G1 humanoid robot, equipped with a Livox Mid-360 LiDAR for 3D environmental perception. The learned policy operates end-to-end, directly mapping raw LiDAR point clouds and proprioceptive state information to low-level motor commands.

    \item \textbf{Simulation Environment:} All policies are trained exclusively in NVIDIA Isaac Sim\cite{mittal2023orbit} and sim2sim in Genesis\cite{Genesis}, a high-fidelity robotics simulator. We employ domain randomization and a structured curriculum, progressively increasing the complexity of obstacles to ensure robust sim-to-real transfer.
\end{itemize}

\subsection{Ablation Study}
 To demonstrate the efficacy of our approach, we compare our proposed method against two well-chosen baselines that ablate key components of our framework. \textbf{Ours (P3O-CBF):} The full proposed policy, trained with P3O using the CBF-based cost functions and the full suite of comfort-oriented rewards. \textbf{PPO-RewardShaping:} Safety is attempted solely by adding a negative reward term that penalizes proximity to obstacles. \textbf{P3O}: A P3O policy is same as P3O-CBF but trained without Comfort-Oriented reward.
    
\vspace{-6pt}
\begin{figure}[h]
    \centering
    \includegraphics[width = 0.48\textwidth]{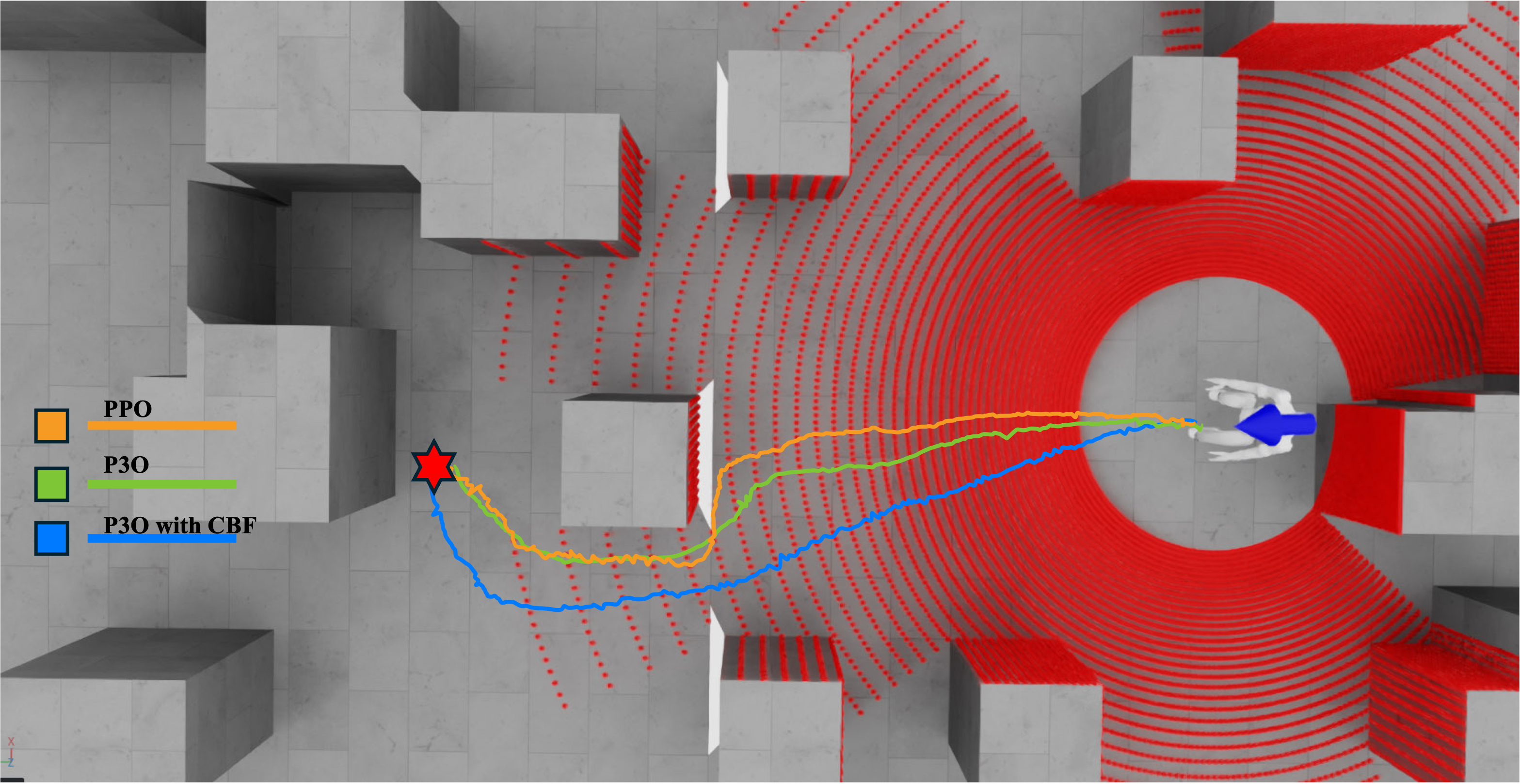}
    \caption{Qualitative comparison of trajectories from the ablation study in a static obstacle course. 
    }
    \label{fig:ablation_study}
\end{figure}
The ablation study highlights the clear advantage of using a principled, dynamics-aware constraint. As shown in Fig. \ref{fig:ablation_study}, the \textbf{PPO(Orange)} policy produces an aggressive trajectory, demonstrating the difficulty of ensuring safety through reward shaping alone. This policy frequently fails by either colliding with obstacles or becoming overly conservative and failing to progress.

The \textbf{P3O(Green)} policy performs better than PPO due to the explicit safety constraint. However, its path is still reactive and close to the obstacles. Because its cost function is only based on distance, it tends to act only when a safety violation is imminent, resulting in less efficient and jerky maneuvers.

In contrast, our proposed \textbf{P3O-CBF(Blue)} policy generates a visibly smoother and safer path. By incorporating a cost function based on CBF principles, which account for the robot's dynamics, the policy learns to anticipate future states. It proactively initiates avoidance maneuvers, resulting in wider turns and a consistently larger safety margin. This leads to the highest success rate and superior comfort metrics, demonstrating that the dynamics-aware cost function is critical for achieving robust, safe, and smooth locomotion.

\subsubsection{Analysis of Safety and Comfort}
We conducted 10 test runs for each policy in a 6-meter-long task space populated with random static obstacles, as depicted in Fig. \ref{fig:ablation_study}. To evaluate performance, we measured the total time the robot spent in two critical regions: the "Unsafe Space," where the distance to an obstacle is less than 0.6m, and the "Uncomfortable Space But Safe," (distance between 0.6m and 1.2m).

\begin{table}[h!]
\centering
\caption{Safety and Comfort Violation Times (in seconds)}
\label{tab:ablation_time}
\resizebox{\columnwidth}{!}{%
\begin{tabular}{@{}lcc@{}}
\toprule
\textbf{Policy} & \textbf{Time in Unsafe Space ($D_{obs} < 0.6$m)} & \textbf{Time in Uncomfortable Space ($0.6 \le D_{obs} < 1.2$m)} \\ \midrule
PPO-RewardShaping & 1.7 s & 3.4 s \\
P3O               & 1.2 s & 3.1 s \\
\textbf{P3O-CBF(Ours)} & \textbf{0.8 s} & \textbf{2.2 s} \\
\bottomrule
\end{tabular}%
}
\end{table}

The results, summarized in Table \ref{tab:ablation_time}, show  that relying on reward shaping alone is insufficient for ensuring robust safety and comfort. The standard P3O policy improved upon this by using an explicit safety constraint, reducing the time spent in the unsafe zone by approximately 30\%. 

Our proposed P3O-CBF method, which includes comfort-oriented rewards, yielded the best performance. It not only minimized the time spent in the unsafe zone (a 53\% reduction compared to PPO) but also significantly reduced the time spent in the caution zone. This indicates that the comfort-oriented rewards successfully encourage the policy to maintain a larger, more socially acceptable distance from obstacles, rather than simply skirting the edge of the hard safety boundary. These quantitative findings, combined with the qualitatively superior trajectories shown in Fig. \ref{fig:ablation_study}, validate the effectiveness of our combined approach for generating locomotion that is both safe and comfortable.

\subsection{Evaluation Scenarios}
The policies are evaluated in a series of challenging simulation scenarios, with the final policy also validated on the physical robot. Each scenario is designed to test a specific capability of our framework.
\begin{figure}[h]
    \centering
    \includegraphics[width = 0.48\textwidth]{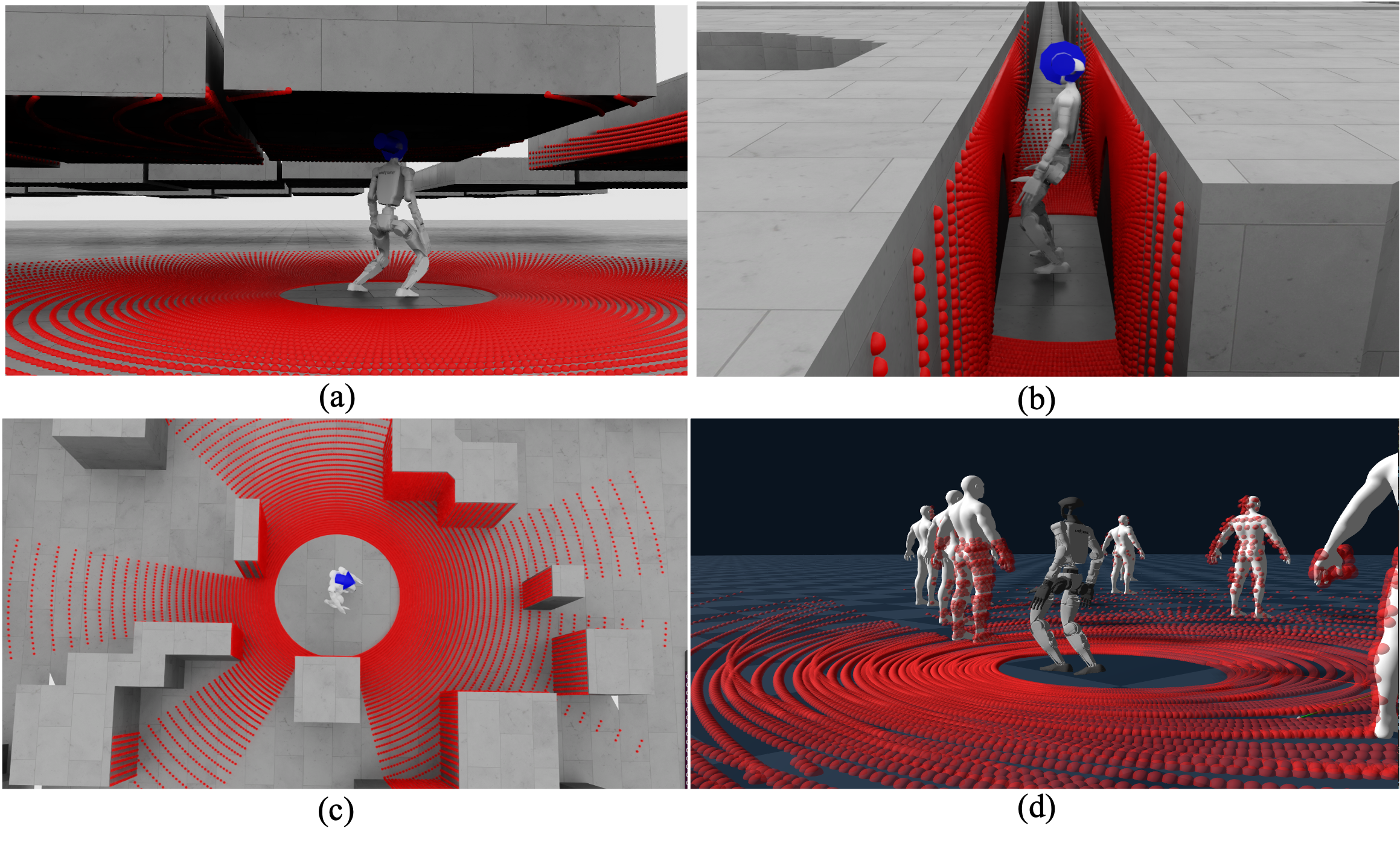}
    \vspace{-24pt}
    \caption{Illustration of the diverse and challenging evaluation scenarios designed. The red concentric circles represent the LiDAR sensor field. (a) \textbf{Suspended Obstacle}: The robot must exhibit full-body awareness to navigate under a low-hanging platform, a task where 2D elevation maps would fail. (b) \textbf{Narrow Passage}: The policy is tested in a confined corridor, requiring precise lateral control to avoid collisions. (c) \textbf{Cluttered Static Course}: The robot navigates a dense field of static pillars, testing its pathfinding in complex environments. (d) \textbf{Dynamic Agents}: The robot demonstrates reactive avoidance while maneuvering among multiple moving humanoids, simulating a crowded, interactive space.}
    \label{fig:evaluation_scenarios}
    \vspace{-6pt}
\end{figure}

The success rates for each policy across the 30 trials per scenario are presented in Table \ref{tab:quantitative_results}. The results clearly demonstrate the superiority of the proposed P3O-CBF framework, especially in the most challenging scenarios.

\begin{table}[h!]
\centering
\caption{Success Rates (\%) of Policies Across Evaluation Scenarios}
\label{tab:quantitative_results}
\resizebox{\columnwidth}{!}{%
\begin{tabular}{@{}l|ccc@{}}
\toprule
\textbf{Scenario} & \textbf{PPO-RewardShaping} & \textbf{P3O} & \textbf{Ours (P3O-CBF)} \\ \midrule
(a) Suspended Obstacle & 20\% & 90\% & 83\% \\
(b) Narrow Passage & 0\% & 33\% & \textbf{60\%} \\
(c) Cluttered Static Course & 93\% & 100\% & \textbf{100\%} \\
(d) Dynamic Agents & 56\% & 70\% & \textbf{86\%} \\
\bottomrule
\end{tabular}%
}
\end{table}

In the \textbf{Suspended Obstacle} scenario, the PPO policy largely fails (20\% success), as its simple reward-based avoidance struggles with the non-standard threat. Both P3O methods perform well, confirming the benefit of explicit constraints for 3D navigation.

The benefit of our CBF-based formulation becomes most apparent in the \textbf{Narrow Passage} and \textbf{Dynamic Agents} scenarios. In the narrow passage, the PPO policy fails entirely (0\%), while our P3O-CBF policy (60\%) nearly doubles the success rate of the P3O with a simple cost (33\%). This suggests that the comfort-oriented reward helps prevent oscillations and over-corrections that lead to collisions in confined spaces. Similarly, when faced with dynamic agents, our method's 86\% success rate significantly outperforms the more reactive P3O (70\%) and PPO (56\%) policies.

\subsection{Physical Tests}
To validate the sim-to-real transfer of our trained policy, we conducted physical experiments on the Unitree G1 humanoid robot in two challenging real-world scenarios. All computations, including perception and policy inference, were performed in real-time on the robot's onboard computer.

\begin{figure}[h!]
    \centering
    \includegraphics[width=\columnwidth]{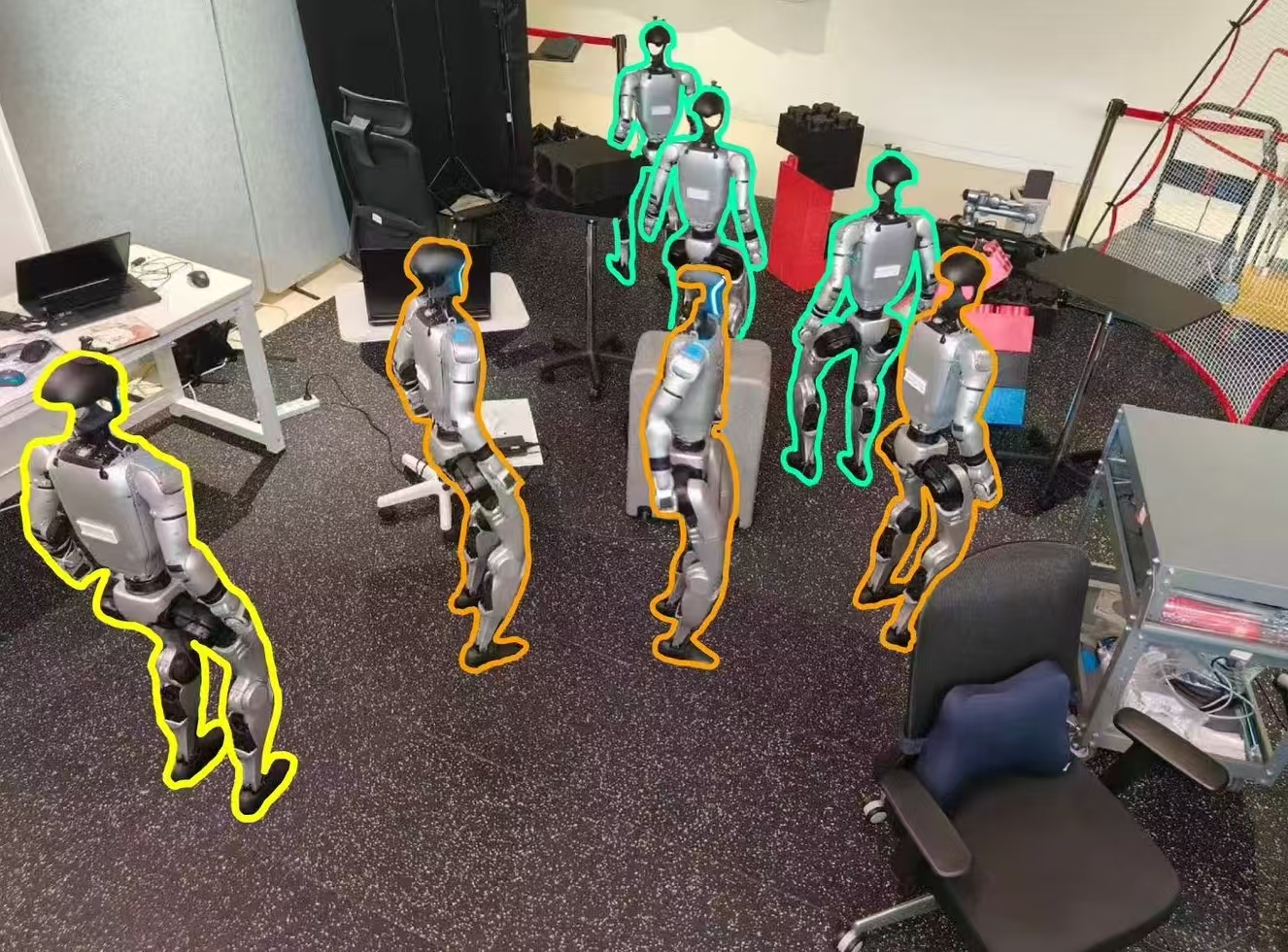}
    \caption{Real-world deployment in a cluttered environment.}
    \label{fig:physical_test_cluttered}
\end{figure}

\begin{figure}[h!]
    \centering
    \includegraphics[width=\columnwidth]{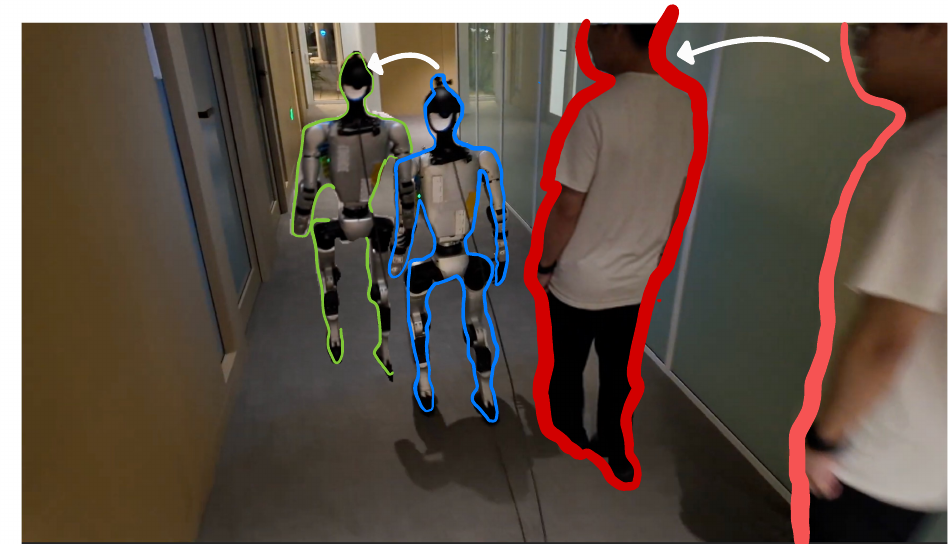}
    \caption{Demonstration of reactive avoidance to a sudden human approach.}
    \label{fig:physical_test_dynamic}
\end{figure}

The first scenario, shown in Fig. \ref{fig:physical_test_cluttered}, tested the policy's core obstacle-avoid capabilities in a cluttered laboratory environment. The robot was commanded to walk through a space populated with various static obstacles. As demonstrated, the policy successfully leveraged its 3D LiDAR perception to maneuver through the complex arrangement without collisions.
The second, more demanding scenario tested the policy's reactivity and safety in a dynamic human-robot interaction context. As shown in Fig. \ref{fig:physical_test_dynamic}, a human agent would suddenly approach the robot from behind. This successful avoidance of an unexpected, dynamic threat underscores the robustness of the learned safety constraints and the effectiveness of the end-to-end framework for real-world deployment.
\section{Conclusions}
\label{sec:conclu}

In this paper, we presented an end-to-end framework for humanoid locomotion that addresses the critical challenges of safety, 3D perception, and human-centric comfort. By leveraging raw LiDAR data, our policy circumvents the limitations of blind and 2D-vision-based controllers, enabling robust navigation in environments with complex 3D obstacles. Our primary contribution is a principled method for integrating the safety guarantees of CBFs into a model-free, constrained reinforcement learning framework. The experimental results validate the efficacy of this approach. Quantitative analysis demonstrated that our proposed P3O-CBF policy significantly outperforms standard PPO and a P3O baseline with a simpler cost function, achieving higher success rates in dynamic and narrow passage scenarios while minimizing time spent in unsafe zones. Furthermore, our ablation studies confirmed that the inclusion of HRI-inspired comfort rewards effectively guides the robot to maintain larger, more socially acceptable distances, resulting in smoother and more predictable motion. Through successful sim-to-real deployment, this work demonstrates a significant step towards developing humanoid robots that are not only dynamically capable but also perceptive, safe, and socially aware, making them better suited for real-world human environments.
\label{headings}

\balance
\bibliographystyle{IEEEtran}
\bibliography{IEEEexample}

\end{document}